\pdfoutput=1

\documentclass[11pt]{article}

\usepackage[final]{acl}

\usepackage{times}
\usepackage{latexsym}

\usepackage[T1]{fontenc}

\usepackage[utf8]{inputenc}

\usepackage{microtype}

\usepackage{inconsolata}

\usepackage{graphicx}

\usepackage{booktabs}
\usepackage{multicol}
\usepackage{multirow}
\usepackage{amssymb}

\usepackage{amsmath}
\usepackage{makecell}

\usepackage{multirow}
\usepackage[normalem]{ulem}
\useunder{\uline}{\ul}{}
\usepackage{amsmath}
\usepackage[ruled,vlined]{algorithm2e}
\usepackage{array} 
\usepackage{makecell}

\usepackage[table]{xcolor}
\usepackage[many,most]{tcolorbox}

\renewcommand\arraystretch{1.32}

\newtcolorbox{mybox}{%
breakable,enhanced,colback=white,colframe=black,left=0.5em,right=0.5em,boxrule=1.0pt}

\newtcolorbox{mybox3}[1]{
  IfBooleanTF={#1}{float*,width=\textwidth}{float},
breakable,enhanced,colbacktitle=white,coltitle=black,colback=white,colframe=black,fonttitle=\bfseries,title=#1,leftupper=0.5em,rightupper=0.5em,boxrule=1.0pt}

\newtcolorbox[auto counter, number within=section]{NewBox}[1]{%
  float*,width=\textwidth,
  colbacktitle=white,coltitle=black,
  colback=white,colframe=black,
  fonttitle=\bfseries,
  title={Box~\thetcbcounter:~#1},
  leftupper=0.5em,rightupper=0.5em,boxrule=1.0pt
}

\usepackage{booktabs}

\title{Improving Medical Calculation of LLMs with Embedded Coding}

\author{Tianshi Ming \\tianshim@andrew.cmu.edu \\
  Carnegie Mellon University\\\And
  Yingying Zhang \\ninzhang@tencent.com \\
  Tencent YouTu Lab
  \\\And
  Xian Wu \\kevinxwu@tencent.com \\
  Tencent YouTu Lab\\}

\begin{document}
\maketitle
\begin{abstract}

Large Language Models (LLMs) perform well on medical examinations and question-answering benchmarks, but remain unreliable on medical calculation tasks that require exact numerical outputs. These calculations support high-stakes decisions such as medication dosing, organ-function assessment, and prognostic scoring, for which even small errors can have serious clinical consequences. We introduce MedCode, a framework that improves medical calculation by training LLMs to generate embedded executable code. Given a clinical context, the model identifies the relevant calculator, extracts its input variables, and produces a script that delegates arithmetic operations to a deterministic interpreter. Executing the script returns the calculated value together with an explanation and the appropriate unit. We construct supervised fine-tuning (SFT) and preference datasets from the MedCalc benchmark and additionally curate a dataset for calculation tasks in Intensive Care Unit (ICU) scenarios. We further propose weighted Direct Preference Optimization (wDPO), which adaptively emphasizes preference pairs that are difficult for the model to distinguish. Experiments with LLaMA3-8B, Qwen2.5-7B, and Mistral-7B show absolute accuracy gains of 20--30 percentage points, demonstrating the effectiveness of embedded code generation for medical calculation.

\end{abstract}

\section{Introduction}

Large Language Models (LLMs), including GPT-4~\cite{gpt4}, have demonstrated strong contextual understanding, attained high performance on medical examinations~\cite{singhal2025toward}, and shown potential in clinical applications such as disease diagnosis~\cite{gao2025leveraging} and treatment planning~\cite{zhou2023survey,stade2024large}. Nevertheless, reliable numerical calculation remains a substantial challenge. Even when a model can recall relevant medical knowledge, it may fail to apply that knowledge correctly in quantitative reasoning~\cite{zhang2025cmedcalc}. This limitation constrains the use of LLMs in clinical settings that require precise numerical outputs.

\begin{figure}[t]
  \centering
  \includegraphics[width=1.0\linewidth]{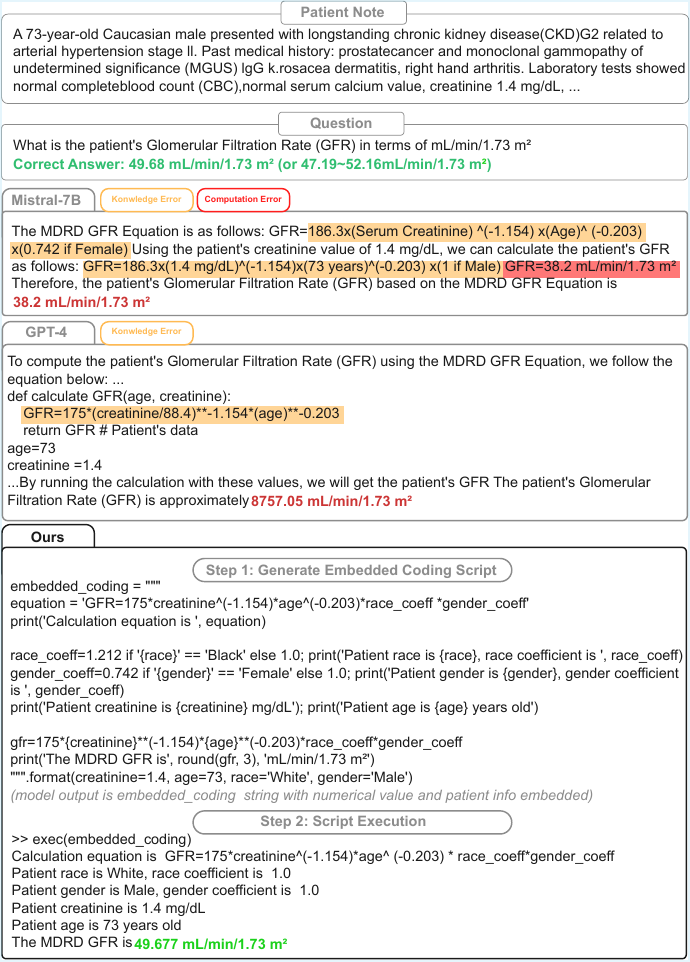}

  \caption{An example medical calculation task. Mistral-7B and GPT-4 produce incorrect answers because of formula-selection or arithmetic errors. MedCode instead generates an embedded code script and executes it with a deterministic interpreter. The execution output contains both the calculation procedure and the numerical answer. This pipeline requires one LLM forward pass for script generation, followed by code execution without an additional model call.}
  \label{fig:case}
\end{figure}

This limitation is especially consequential in the Intensive Care Unit (ICU), where clinical decisions frequently depend on timely and accurate calculations. For example, computing the Glomerular Filtration Rate (GFR), an indicator of renal function, requires selecting the appropriate equation and extracting variables such as serum creatinine and age from a patient note. As illustrated in Figure~\ref{fig:case}, GPT-4 omits coefficients required by the specified calculator, while Mistral-7B exhibits both formula-selection and arithmetic errors. Such failures may propagate to downstream decisions, including medication dosing and prognostic assessment. Consistent with the findings of MedCalc-Bench~\cite{MedCalc}, these examples highlight the need for methods that improve numerical reliability rather than relying on direct text generation alone.

Our analysis suggests that many failures arise after the relevant information has been identified: models may extract the required variables correctly but make errors in multi-step arithmetic or in the constants and conditions of a medical formula. This observation motivates separating contextual interpretation from numerical computation. The LLM can parse the clinical context and instantiate the required calculation, while a deterministic interpreter performs the arithmetic.

Motivated by this separation, we propose MedCode. Rather than directly generating a numerical answer, the model is fine-tuned to produce an executable embedded code script that combines calculation logic with textual output. As shown in Figure~\ref{fig:case}, arithmetic operations are expressed as code, while explanations and units are emitted through \texttt{print()} statements. A single script execution therefore returns both the numerical result and its accompanying explanation. MedCode requires one LLM decoding pass followed by deterministic execution, avoiding the repeated model calls used in multi-stage function-calling pipelines~\cite{shen2024llm}.

To train models for this generation format, we convert MedCalc-Bench examples into executable scripts for supervised fine-tuning (SFT). We then collect erroneous scripts produced by the fine-tuned model and pair each with its correct reference script, yielding preference data without requiring separate pairwise annotation. We additionally construct a clinician-authored synthetic dataset for ICU scenarios, where a model must extract, filter, and count longitudinal clinical measurements.

Building on these datasets, we introduce weighted Direct Preference Optimization (wDPO). Standard preference optimization treats all response pairs uniformly, although pairs that differ only in an extracted argument, coefficient, or unit conversion can be more difficult to distinguish than pairs containing an obvious structural error. wDPO assigns adaptive weights to the preference loss according to the probability gap between the preferred and dispreferred scripts, thereby emphasizing ambiguous pairs during training.

The contributions of this paper are as follows:
\begin{itemize}
    \item We propose MedCode, an embedded code generation framework that improves medical calculation accuracy by 20--30 absolute percentage points.
    \item We introduce an error-driven procedure for constructing preference pairs and propose wDPO to optimize a code-generating model with difficulty-aware preference weights.
    \item We evaluate MedCode on the public MedCalc benchmark and a newly constructed ICU calculation dataset. The datasets and code will be released upon publication.
\end{itemize}

\section{Related Work}
\subsection{Medical Large Language Model}

Medical large language models (Med-LLMs) have been developed for a broad range of clinical and biomedical tasks~\cite{huatuo-2,med-gemini,taiyi,ophglm,med-palm}. However, medical metric calculation requires a distinct combination of domain knowledge, information extraction, and exact numerical computation. Common benchmarks, including PubMedQA~\cite{pubmedqa}, MedQA~\cite{medqa}, MedMCQA~\cite{medmcqa}, and MMLU~\cite{mmlu}, primarily use multiple-choice questions and therefore do not directly evaluate open-ended calculation. MedCalc-Bench~\cite{MedCalc} instead requires models to produce answers for equation-based and rule-based medical calculators without predefined options, and reports low accuracy for both open- and closed-source models. MedCode addresses this setting by combining domain-specific supervised fine-tuning with preference learning for executable medical calculation.

\subsection{Coding Generation}

Code-assisted reasoning has been widely studied as a means of reducing computational errors in LLM outputs~\cite{code-hallu}. Verification-based methods such as CSV~\cite{csv} use an interpreter to check generated answers, whereas ToRA~\cite{tora} interleaves natural-language reasoning and code execution. Because execution results are fed back into subsequent reasoning steps, these methods may require multiple model calls. MedCode instead trains the model to emit a complete executable script in one decoding pass, followed by deterministic execution.

Program-aided methods directly generate code whose execution supplies the answer. PAL~\cite{pal} and PoT~\cite{pot} combine intermediate reasoning with program generation, while Chain of Code~\cite{coc} simulates line-by-line code execution. Medical calculation differs from general-purpose program synthesis because the generated program must encode the correct clinical calculator, units, coefficients, and input variables. We therefore construct domain-specific code training data rather than relying exclusively on inference-time prompting. Our experiments show that models fine-tuned on these data can outperform substantially larger models that receive only prompt-based code demonstrations.

Agent-based systems such as AgentMD~\cite{agentmd}, EHRAgent~\cite{ehragent}, TOOLMAKER~\cite{toolmaker}, and REFLECTOOL~\cite{reflectool} coordinate tool selection, execution, and verification across multiple inference stages. These frameworks provide broad tool-use capabilities but incur additional model calls and are not designed specifically for medical metric calculation. By contrast, MedCode encodes calculator knowledge in its training data and generates the complete calculation program in a single model pass. The resulting script is then executed without further agent orchestration.

\section{Dataset Generation}

\begin{table*}[!t]
    \centering
    \fontsize{9}{9.5}\selectfont

    \setcellgapes{0.7pt}\makegapedcells
    \setlength{\aboverulesep}{0.2pt}
    \setlength{\belowrulesep}{0.2pt}
    \setlength{\tabcolsep}{3pt}

    \begin{tabular}{ll}
        \toprule

        \multicolumn{2}{l}{
            \textbf{> Equation-based Calculator Example}:
            Modification of Diet in Renal Disease Glomerular Filtration Rate (MDRD GFR)
        } \\

        \midrule

        \makecell[l]{\textbf{Patient Note}} &
        \makecell[l]{
            A 73-year-old Caucasian male presented with longstanding chronic kidney disease \\
            (CKD) G2 related to arterial hypertension stage II. Past medical history: prostate \\
            cancer and monoclonal gammopathy of undetermined significance (MGUS) IgG k, \\
            rosacea dermatitis, right hand arthritis. ...
        } \\

        \cmidrule(lr){2-2}

        \makecell[l]{\textbf{Question}} &
        \makecell[l]{
            What is the patient's Glomerular Filtration Rate (GFR) in terms of
            mL/min/1.73 m\textsuperscript{2}?
        } \\

        \cmidrule(lr){2-2}

        \makecell[l]{\textbf{Generated Code}} &
        \makecell[l]{
            \begingroup
            \setcellgapes{0.35pt}\makegapedcells
            \renewcommand{\arraystretch}{0.9}
            \begin{tabular}[t]{@{}l@{}}
                \# GFR = 175 * creatinine\^{}(-1.154) * age\^{}(-0.203) * race\_coefficient * gender\_coefficient \\
                race\_coefficient = 1.212 if 'Black' == 'Black' else 1.0 \\
                gender\_coefficient = 0.742 if 'Male' == 'Female' else 1.0 \\
                gfr = 175 * 10.6**(-1.154) * 49**(-0.203) * race\_coefficient * gender\_coefficient \\
                gfr = round(gfr, 3) \\
                print('The MDRD GFR is ', gfr, ' mL/min/1.73 m\textsuperscript{2}')
            \end{tabular}
            \endgroup
        } \\

        \cmidrule(lr){2-2}

        \makecell[l]{\textbf{Code Execution Result}} &
        \makecell[l]{
            The MDRD GFR is 49.677 mL/min/1.73 m\textsuperscript{2}
        } \\

        \midrule

        \multicolumn{2}{l}{
            \textbf{> Rule-based Calculator Example}:
            Centor Score (Modified/McIsaac) for Strep Pharyngitis
        } \\

        \midrule

        \makecell[l]{\textbf{Patient Note}} &
        \makecell[l]{
            A 15-year-old female presents to your office for evaluation of a painful neck mass. \\
            The family reports the patient has had a midline neck mass for many years, but had \\
            become painful over the past week with erythema at the overlying skin. ...
        } \\

        \cmidrule(lr){2-2}

        \makecell[l]{\textbf{Question}} &
        \makecell[l]{
            What is the patient's Centor Score?
        } \\

        \cmidrule(lr){2-2}

        \makecell[l]{\textbf{Generated Code}} &
        \makecell[l]{
            \begingroup
            \setcellgapes{0.35pt}\makegapedcells
            \renewcommand{\arraystretch}{0.9}
            \begin{tabular}[t]{@{}l@{}}
                \# Centor Score (Modified/McIsaac) criteria: age, temperature, ... \\
                \# age thresholds are: (45, 'inf'), (3, 15) \\
                ... \\
                \# The patient has tender/swollen anterior cervical lymph nodes or not \\
                score = 0 \\
                score += -1 * (45 <= 15 < float('inf')) + 1 * (3 <= 15 < 15) \\
                ... \\
                has\_exudate\_swelling\_tonsils = False \\
                score += 1 if has\_exudate\_swelling\_tonsils else 0 \\
                print('The Centor Score (Modified/McIsaac) for Strep Pharyngitis is', score)
            \end{tabular}
            \endgroup
        } \\

        \cmidrule(lr){2-2}

        \makecell[l]{\textbf{Code Execution Result}} &
        \makecell[l]{
            Centor Score (Modified/McIsaac) for Strep Pharyngitis is 2
        } \\

        \bottomrule
    \end{tabular}

    \caption{Examples of embedded code generation for equation-based and rule-based medical calculation tasks.}
    \label{tab:code-example}
\end{table*}
\begin{figure*}[t]
  \centering
  \includegraphics[width=1.0\linewidth]{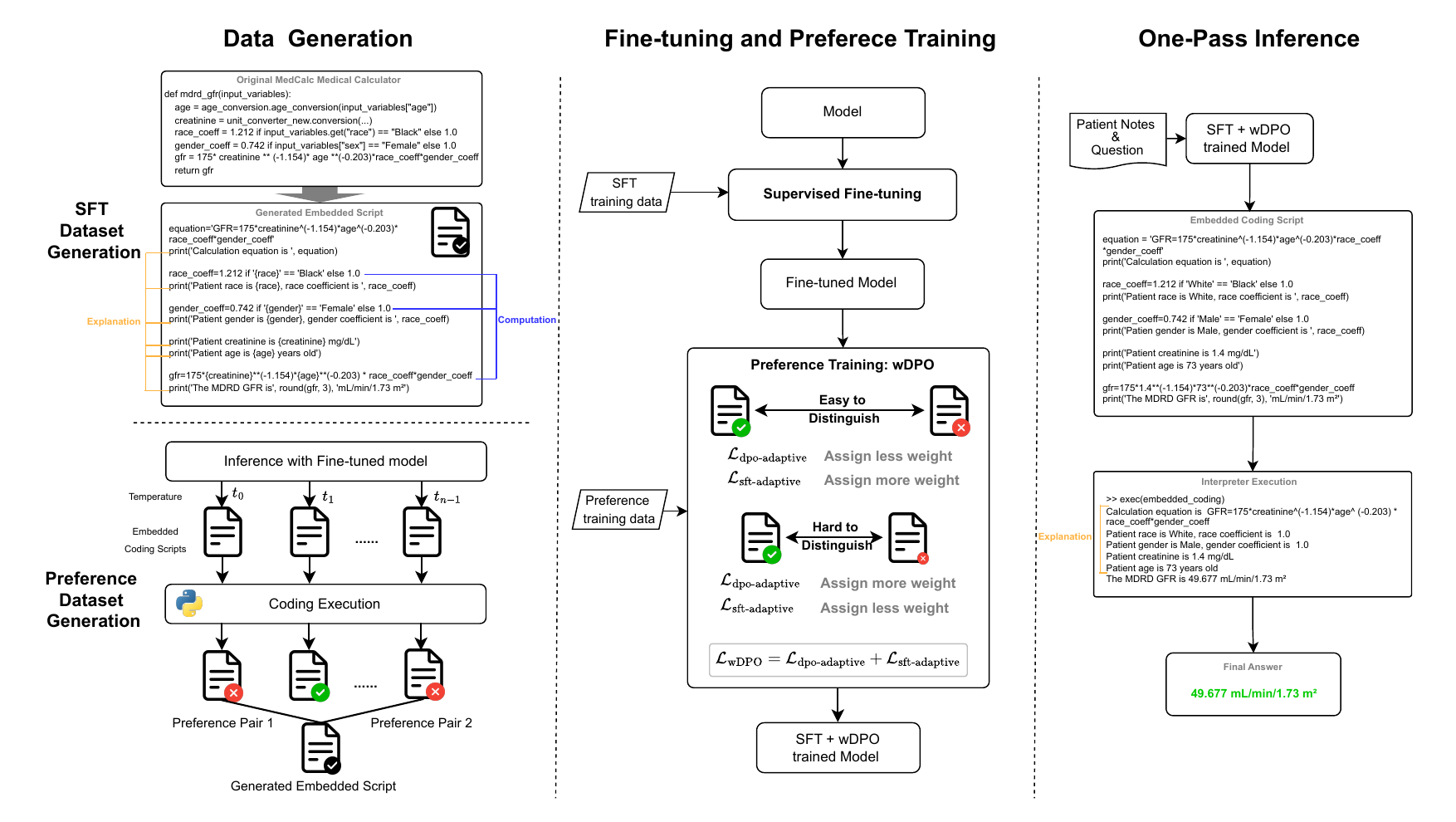}

  \caption{Overview of MedCode. \textbf{1) Data generation}: We convert MedCalc calculators into an embedded code dataset for SFT and pair erroneous scripts produced by the fine-tuned model with reference scripts to construct preference data. \textbf{2) Fine-tuning and preference training}: We first perform SFT and then apply wDPO, which adaptively weights the preference objective. \textbf{3) One-pass inference}: The model generates a script containing values extracted from the patient note, and a deterministic interpreter executes the script to produce the final answer and calculation explanation.}
  \label{fig:main}
\end{figure*}


\subsection{Embedded Coding Dataset Generation based on MedCalc}
\label{sec:sftdata}

MedCalc-Bench~\cite{MedCalc} comprises 55 commonly used medical calculation tasks collected from online medical calculators\footnote{https://www.mdcalc.com/\#Popular}. Each example contains a patient note, a calculation question, and a manually reviewed reference answer (Figure~\ref{fig:main}). Solving an example requires selecting the relevant calculator and applying it to clinical variables described in the note.

We use code as an intermediate representation and delegate numerical operations to an executable program. Based on the MedCalc calculators, we construct an embedded code dataset with three components. \textbf{(1) Medical knowledge}: code comments specify the calculator logic and formula. \textbf{(2) Embedded computation}: a compact script instantiates the formula with values extracted from the note and applies explicit conversion factors when units differ. \textbf{(3) Output generation}: the script prints the result together with an explanation and the corresponding unit. Appendix~\ref{app:eq-implmentations} details the script designs.

\subsection{Preference Dataset Generation Based on MedCalc}

We first fine-tune the model on the embedded code dataset. Although SFT substantially improves accuracy, the resulting model can still extract an incorrect argument or generate an erroneous expression. Such failures often differ from the reference script by only a coefficient, variable, or implementation detail. We therefore introduce preference learning to train the model to distinguish correct scripts from these plausible but incorrect alternatives.

Because explicitly annotated preference data for medical calculation code are scarce, we construct preference pairs from errors produced by the SFT model, as illustrated in Figure~\ref{fig:main}. The procedure consists of the following two steps.

\paragraph{Inference on training data} We sample $k$ instances from the SFT dataset described in Section~\ref{sec:sftdata}, preserving the original distribution over equation-based and rule-based tasks. A new subset is sampled for each inference run. The fine-tuned model then generates scripts for these instances at temperatures from 0 to 1 in increments of 0.1. Varying the temperature increases output diversity and exposes a broader set of failure modes. Appendix~\ref{app:prompt} provides the sampling and inference settings.

\paragraph{Pairing error scripts with correct scripts}

We execute the generated scripts and retain outputs that are incorrect or fail to execute. Each erroneous script is paired with the reference implementation for the same input, yielding a dispreferred response and a preferred response. This error-driven construction produces preference pairs without requiring an independently annotated preference corpus.



\subsection{ICU-Dataset}
\label{app:icu-data}

MedCalc-Bench evaluates calculators applied to a single summarized patient state. ICU workflows, by contrast, often require operations over measurements recorded at multiple time points. The resulting tasks combine information extraction, threshold-based filtering, and counting over longitudinal Electronic Health Records (EHRs), placing simultaneous demands on contextual understanding and numerical reliability.

We therefore construct a dataset for medical statistical calculation in ICU scenarios. Each instance provides longitudinal EHR entries and a threshold-based criterion; the model must identify the entries satisfying the criterion and report their number. We evaluate two aspects of performance: \textbf{(1) extraction accuracy}, the proportion of relevant records that are correctly extracted, and \textbf{(2) frequency counting}, the accuracy of the final count. The dataset is written and verified by experienced clinicians.

\paragraph{Data Format} 

The ICU dataset is manually constructed and verified by experienced clinicians. Each synthetic example contains basic patient information and a set of timestamped EHR entries, including metric names and values. An SFT instance consists of an instruction, the EHR entries, and a compact reference script. Given a threshold or interval, the model must identify and count the entries satisfying the specified condition. The dataset is written in simplified Chinese and contains 5,000 training examples and 1,000 test examples. Appendix~\ref{box:icu-dataset-new} presents a translated example.

\paragraph{Data Augmentation} 

We augment the dataset by recombining existing entries and sampling new values within clinically plausible ranges. Table~\ref{tab:icu-metrics} summarizes the clinical categories, metrics, units, and value ranges. The resulting dataset covers 7 clinical categories and 16 medical metrics.

\section{Preference Training Strategy}

Supervised fine-tuning substantially improves calculation accuracy, but the model continues to produce subtle script errors, including incorrect coefficients and unit conversions. We therefore apply preference optimization to distinguish reference scripts from erroneous alternatives.

\subsection{Weighted Direct Preference Optimization}
\label{sec:wdpo}

Preference learning trains the model to distinguish a preferred script $y_w$ from a dispreferred script $y_l$ for input $x$, as illustrated in Figure~\ref{fig:main}. Direct Preference Optimization (DPO)~\cite{dpo} compares the likelihood ratios of these responses under the policy $\pi_\theta$ and a fixed reference model $\pi_{\mathrm{ref}}$. Its objective is given in Equation~\ref{eq:dpo}.
\begin{equation}
\label{eq:dpo}
\begin{footnotesize}
\begin{aligned}
    &\mathcal{L}_{\mathrm{dpo}}\left(\pi_\theta ; \pi_{\text {ref}}\right) \\
  &=-\mathbb{E}_{(x, y)\!\sim\mathcal{D}}\!\left[\log\sigma\!\left(\!\beta\log\!\frac{\pi_{\theta}(y_w|x)}{\pi_{\text{ref}}(y_w|x)} - \beta\log\!\frac{\pi_{\theta}(y_l|x)}{\pi_{\text{ref}}(y_l|x)}\!\right)\right]  \\
  &=-\mathbb{E}_{(x, y)\sim\mathcal{D}}\left[\log \sigma\left(\beta \cdot \hat r(x, y_w)  - \beta \cdot \hat r(x, y_l) \right)\right]. 
\end{aligned}
\end{footnotesize}
\end{equation}
where $\hat r(x,y_w)$ and $\hat r(x,y_l)$ denote the implicit rewards assigned to the preferred and dispreferred scripts, respectively. Standard DPO assigns equal importance to every preference pair, although the pairs vary in difficulty. For example, a missing computation step creates a conspicuous structural difference, whereas an incorrect unit-conversion factor may differ from the reference by only a single constant. Appendix~\ref{app:case-study-error} provides examples of these minor and substantial errors.

We address this variation with an adaptive preference objective that weights each pair according to how clearly the current model separates its two responses. Let $w\!\left(\pi_{\theta}(y_w|x),\pi_{\theta}(y_l|x)\right)$ denote this weight. Pairs with similar model probabilities receive larger weights, whereas pairs with a clear probability gap receive smaller weights. The resulting objective $\mathcal{L}_{\mathrm{dpo-adaptive}}$ is defined in Equation~\ref{eq:wdpo}.
\begin{equation}
\label{eq:wdpo}
\begin{footnotesize}
\begin{aligned}
    &\mathcal{L}_{\mathrm{dpo-adaptive}}\left(\pi_\theta ; \pi_{\text {ref }}\right)\\
    &=-\mathbb{E}_{(x, y)\sim\mathcal{D}} w\left(\pi_{\theta}(y_w|x), \pi_{\theta}(y_l|x)\right)\left[\log\!\sigma\left(\beta \Delta \hat r \right)\right].
    \end{aligned}
\end{footnotesize}
\end{equation}
where $\Delta \hat r=\hat r(x,y_w)-\hat r(x,y_l)$. We use the normalized difference between $\pi_{\theta}(y_w|x)$ and $\pi_{\theta}(y_l|x)$ as the model-dependent measure of pairwise separation. The adaptive weight is defined as follows:
\begin{equation}
\begin{footnotesize}
\begin{aligned}
    &w(\pi_{\theta}(y_w|x),\pi_{\theta}(y_l|x))
    = 1 - \left|\frac{\pi_{\theta}(y_w|x) - \pi_{\theta}(y_l|x)}{|\pi_{\theta}(y_w|x)|+|\pi_{\theta}(y_l|x)|}\right|.
\end{aligned}
\end{footnotesize}
\end{equation}

The denominator applies an $L_1$ normalization to keep the probability difference on a common scale. Taking the absolute value makes the weight depend on the magnitude of separation, including cases in which the current model ranks $y_l$ above $y_w$. The probabilities used to compute $w$ are detached from the computation graph, so gradients do not propagate through the weight. We recompute these probabilities after each training epoch.

The adaptive DPO term emphasizes pairs whose responses are difficult to separate. Conversely, pairs with a large probability gap receive a smaller preference weight and may contribute less to training. To retain direct supervision from their preferred responses, we add an SFT objective with the complementary coefficient $1-w$. The adaptive SFT term is defined in Equation~\ref{eq:sftreg}.
\begin{equation}
\label{eq:sftreg}
\begin{footnotesize}
\begin{aligned}
    &\mathcal{L}_{\mathrm{sft-adaptive}}\left(\pi_\theta ; \pi_{\text {ref }}\right)\\
    &=-\mathbb{E}_{(x, y)\sim\mathcal{D}} (1\!-\!w\left(\pi_{\theta}(y_w|x), \pi_{\theta}(y_l|x))\right)\left[\log\!\pi_{\theta}(y_w|x)\right].
    \end{aligned}
\end{footnotesize}
\end{equation}

When $w$ is small, $\mathcal{L}_{\mathrm{sft-adaptive}}$ places greater weight on maximizing the likelihood of the preferred response. We define the complete wDPO objective as the sum of the adaptive DPO and adaptive SFT terms in Equation~\ref{eq:loss-full}.
\begin{equation}
\label{eq:loss-full}
\begin{footnotesize}
\begin{aligned}
    &\mathcal{L}_{\mathrm{wDPO}}\left(\pi_\theta ; \pi_{\text {ref }}\right) =  \\ &\mathcal{L}_{\mathrm{dpo-adaptive}}\left(\pi_\theta ; \pi_{\text {ref }}\right) + \mathcal{L}_{\mathrm{sft-adaptive}}\left(\pi_\theta ; \pi_{\text {ref }}\right).
\end{aligned}
\end{footnotesize}
\end{equation}

\begin{table*}[h]
    \centering
    \small
    \renewcommand\arraystretch{1.0}
    \begin{tabular}{lcccccccc}
        \toprule
        \textbf{Model} &
        \multicolumn{4}{c}{\textbf{Equation-based}} &
        \multicolumn{3}{c}{\textbf{Rule-based}} &
        \textbf{Avg.} \\
        \cmidrule(lr){2-5} \cmidrule(lr){6-8}
        & Lab & Phys. & Date & Dosage & Risk & Sev. & Diag & \\
        \midrule

        \multicolumn{9}{c}{\textbf{Zero-shot}} \\
        \midrule
        LLaMA3-8B&10.70&19.17&3.33&5.00&12.50&8.75&25.00&13.09 \\         
        Mistral-7B&10.70&18.33&3.33&0.00&4.58&3.75&13.33&9.8 \\    
        GPT-3.5&17.13&35.00&13.33&5.00&12.92&6.25&18.33&18.82 \\
        GPT-4&14.37&34.58&38.33&15.00&14.58&15.00&20.00&20.82 \\    
        
        \midrule
        \multicolumn{9}{c}{\textbf{Zero-shot CoT}} \\
        \midrule
        LLaMA3-8B&16.51&25.00&1.67&7.50&11.25&\textbf{13.75}&26.67&16.43 \\     
        Mistral-7B&10.09&14.58&1.67&0.00&9.58&7.50&25.00&10.79 \\ 
        GPT-3.5&20.49&45.00&11.67&17.50&13.33&10.00&31.67&23.69 \\
        GPT-4&26.30&71.25&\textbf{48.33}&40.00&27.50&15.00&28.33&37.92 \\ 

        \midrule
        \multicolumn{9}{c}{\textbf{One-shot CoT}} \\
        \midrule
        LLaMA3-8B&34.86&35.42&3.33&2.50&20.00&11.25&41.67&27.13\\
        Mistral-7B&11.01&30.42&6.67&0.00&16.25&6.25&18.33&16.05 \\
        GPT-3.5&30.89&59.17&41.67&15.00&23.33&17.50&35.00&34.86 \\
        GPT-4&51.68&77.50&46.67&37.50&\textbf{33.75}&27.50&53.33&50.91 \\ 

        \midrule
        \multicolumn{9}{c}{\textbf{SFT Direct}} \\
        \midrule
        LLaMA3-8B&19.27&85.00&33.33&28.57&11.25&10.00&41.67&33.37\\
        Mistral-7B&19.27&91.25&31.67&17.50&18.75&15.00&43.33&37.34 \\

        \midrule
        \multicolumn{9}{c}{\textbf{SFT Code}} \\
        \midrule
        LLaMA3-8B&66.26&92.53&26.67&42.50&23.24&15.00&50.00&54.39 \\
        Mistral-7B&70.55&\textbf{99.59}&33.33&45.00&23.24&\textbf{18.75}&46.67&57.92 \\
        GPT-3.5*&32.21&34.02&36.67&15.00&4.15&7.50&13.33&22.81 \\
        GPT-4*&57.67&90.46&36.67&45.00&23.24&11.25&\textbf{97.50}&52.86 \\

        \midrule
        \multicolumn{9}{c}{\textbf{SFT Code+wDPO}} \\
        \midrule
        LLaMA3-8B&71.78&83.82&33.33&\textbf{65.00}&22.82&10.00&58.33&55.34 \\
        Mistral-7B&\textbf{76.38}&\textbf{99.59}&33.33&45.00&23.24&\textbf{18.75}&46.67&\textbf{59.73} \\
        
        \bottomrule
    \end{tabular}

    \caption{Results on MedCalc-Bench. We evaluate LLaMA3-8B-Instruct, Mistral-7B-Instruct-v0.2, GPT-3.5-Turbo, and GPT-4. CoT denotes Chain-of-Thought prompting. Zero-shot, zero-shot CoT, and one-shot CoT results are taken from MedCalc-Bench~\citep{MedCalc}. *For code generation with GPT-3.5 and GPT-4, we use one in-context example.}
    \label{tb:main}
\end{table*}

\section{Experiments}
\subsection{Experiment Settings}

We evaluate LLaMA3-8B-Instruct~\cite{llama3}, Mistral-7B-Instruct~\cite{mistral}, Qwen2.5-7B-Instruct~\cite{qwen2.5}, and Baichuan2-7B-Chat~\cite{baichuan2}; GPT-3.5 and GPT-4 are included as prompted baselines. For the inference-time baselines, we follow the prompts and inference settings released with MedCalc-Bench~\cite{MedCalc}. We implement SFT and preference optimization with LLaMA-Factory~\cite{llamafactory}. Appendix~\ref{app:prompt} reports the complete training, inference, and prompt configurations.

\subsection{Main Results}

We first evaluate MedCode on MedCalc-Bench, which contains four equation-based and three rule-based task categories. Its training split comprises 9,766 examples with patient notes, questions, reference answers, and the medical entities required by each calculator. Answers include continuous values, discrete scores, dates, and time intervals. We compare inference-time prompting (zero-shot, one-shot, and Chain-of-Thought) with three training-based variants: direct-answer SFT (SFT Direct), embedded-code SFT (SFT Code), and SFT followed by wDPO (SFT Code+wDPO). Table~\ref{tb:main} reports the results.

Training-based methods substantially outperform inference-time prompting for the same backbone. For LLaMA3-8B, average accuracy increases from 27.13 with one-shot CoT to 33.37 with SFT Direct and 54.39 with SFT Code, corresponding to gains of 6.24 and 27.26 percentage points. For Mistral-7B, the corresponding scores are 16.05, 37.34, and 57.92, yielding gains of 21.29 and 41.87 points. Adding wDPO further raises average accuracy by 0.95 points for LLaMA3-8B and 1.81 points for Mistral-7B. Both SFT-Code models also exceed the reported GPT-4 averages.

The gains from wDPO are concentrated in specific task categories. Relative to SFT Code, it improves laboratory-task accuracy by 5.52 points for LLaMA3-8B and 5.83 points for Mistral-7B. On LLaMA3-8B, the dosage and diagnosis categories improve by 22.50 and 8.33 points, respectively, although several other categories remain unchanged or decrease. These results indicate that approximately 1,500 mined preference pairs can improve selected fine-grained behaviors, while also showing that the benefit is not uniform across tasks.

\subsection{Evaluation on ICU dataset}

We next evaluate MedCode on the ICU dataset (Table~\ref{tb:icu}). For LLaMA3-8B, replacing direct-answer SFT with embedded-code SFT increases extraction accuracy from 52.39 to 82.10 and frequency-count accuracy from 75.10 to 97.60, gains of 29.71 and 22.50 percentage points. wDPO provides additional gains to 83.20 and 98.00. Qwen2.5-7B shows the same overall pattern, with a particularly large improvement in frequency counting. These results support the use of executable code for threshold-based extraction and aggregation.


\begin{table}[th]
    \centering
    \small
    \renewcommand\arraystretch{1.0}
    \begin{tabular}{clccc}
        \toprule
        \multicolumn{1}{c}{\textbf{Model}}&
        \multicolumn{1}{c}{\textbf{Methods}}&
        \multicolumn{1}{c}{\textbf{extr.}}&
        \multicolumn{1}{c}{\textbf{fcnt.}} \\
        \midrule
        LLaMA3-8B
        &One-shot Direct & 39.22 &  0.0 \\
        &SFT-Direct & 52.39 &  75.10 \\
        &SFT-Code & 82.10 &  97.60 \\
        &SFT-Code-wDPO & \textbf{83.20} &  \textbf{98.00} \\
        \midrule
        Qwen2.5-7B
        &One-shot Direct & 47.90 &  0.0 \\
        &SFT-Direct & 56.70 &  77.90 \\
        &SFT-Code & 59.40 &  97.00 \\
        &SFT-Code-wDPO & \textbf{62.10} &  \textbf{97.50} \\
        
        \bottomrule
    \end{tabular}

    \caption{Results on the ICU dataset. ``extr.' denotes extraction accuracy and ``fcnt.' denotes frequency-count accuracy. We evaluate LLaMA3-8B-Instruct and Qwen2.5-7B-Instruct.}
    \label{tb:icu}
\end{table}

We additionally compare MedCode with reasoning-focused LLMs on the ICU dataset (Table~\ref{tab:icu_reasoning}). LLaMA3-8B-Instruct trained with SFT and wDPO reaches 83.20 extraction accuracy and 98.00 frequency-count accuracy, compared with 85.87 and 99.53 for DeepSeek-R1. Although DeepSeek-R1 and OpenAI-o3 obtain higher scores, the smaller fine-tuned model remains competitive on this task. This comparison is relevant to clinical deployment, where restrictions on sending data to external APIs and limited local compute can constrain the use of proprietary services or very large models.


\begin{table}[t]
\centering
\small
\renewcommand\arraystretch{1.0}
\setlength{\tabcolsep}{5pt}
\begin{tabular}{lcc}
\toprule
\textbf{Model} & \textbf{extr.} & \textbf{fcnt.} \\
\midrule
LLaMA3-8B-Instruct (wDPO) & 83.20 & 98.00 \\
Qwen2.5-7B-Instruct (wDPO) & 62.10 & 97.50 \\
Baichuan2-7B-Chat (wDPO) & 79.13 & 96.54 \\
DeepSeek-R1 & 85.87 & 99.53 \\
OpenAI-o3 & 90.10 & 99.10 \\
\bottomrule
\end{tabular}

\caption{Comparison with reasoning-focused models on the ICU dataset. ``extr.' denotes extraction accuracy and ``fcnt.' denotes frequency-count accuracy.}
\label{tab:icu_reasoning}
\end{table}

\subsection{Ablation Study on Adaptive SFT}

We study the complementary SFT term introduced in Section~\ref{sec:wdpo} by comparing DPO and wDPO with fixed or adaptive regularization (Table~\ref{tb:reg}). Adding a fixed SFT term improves both DPO and wDPO over their unregularized variants. For wDPO, adaptive regularization achieves the highest equation-based accuracy (72.26) and overall average (55.34), while a fixed coefficient of 1.0 yields the highest rule-based score (29.40). Thus, adaptive regularization provides the best aggregate result, but its advantage is not uniform across task types.

\begin{table*}[t]
\centering
\small
\renewcommand\arraystretch{1.0}
\setlength{\tabcolsep}{1.05mm} 
\begin{tabular}{@{}lcccccccc@{}}
\toprule
\textbf{Strategy} & \textbf{Lab} & \textbf{Phys.} & \textbf{Date} & \textbf{Dos.} & \textbf{Risk} & \textbf{Sev.} & \textbf{Diag.} & \textbf{Avg.} \\
\midrule
Zero-shot          & 10.70 & 19.17 & 3.33 & 5.00 & 12.50 & 8.75 & 25.00 & 13.09 \\
Zero-shot CoT      & 16.51 & 25.00 & 1.67 & 7.50 & 11.25 & \textbf{13.75} & 26.67 & 16.43 \\
One-shot CoT       & \textbf{34.86} & 35.42 & 3.33 & 2.50 & \textbf{20.00} & 11.25 & 41.67 & 27.13 \\
Zero-shot SFT Code & 19.27 & 85.00 & 33.33 & 28.57 & 11.25 & 10.00 & 41.67 & 33.37 \\
One-shot SFT Code  & 20.38 & 89.85 & 35.25 & 30.24 & 11.92 & 10.59 & 44.08 & 34.56 \\
3-shot SFT Code    & 21.49 & \textbf{94.72} & \textbf{37.17} & \textbf{31.89} & 12.55 & 11.15 & \textbf{46.49} & \textbf{36.77} \\
\bottomrule
\end{tabular}
\caption{Comparison of zero and few-shot strategies.}
\label{tab:zero-few-shot-ablation-full}
\end{table*}

\begin{table*}[t]
\centering
\small
\renewcommand\arraystretch{0.75}
\setlength{\tabcolsep}{0.75mm} 
\begin{tabular}{@{}lcccccccc@{}}
\toprule
\textbf{Method} & \textbf{Lab} & \textbf{Phys.} & \textbf{Date} & \textbf{Dosage} & \textbf{Risk} & \textbf{Sev.} & \textbf{Diag.} & \textbf{Avg.} \\
\midrule
rDPO~\citep{rdpo} & 68.45 & 79.20 & 30.12 & 59.33 & \textbf{25.10} & \textbf{12.55} & 51.33 & 50.84 \\
ROPO~\citep{ropo} & 70.80 & 82.50 & \textbf{34.00} & 63.75 & 23.90 & 11.30 & 53.45 & 54.25 \\
WPO~\citep{wpo}   & 69.20 & 81.00 & 31.45 & 61.25 & 24.70 & 11.00 & 47.70 & 52.33 \\
wDPO (ours)       & \textbf{71.78} & \textbf{83.82} & 33.33 & \textbf{65.00} & 22.82 & 10.00 & \textbf{58.33} & \textbf{55.34} \\
\bottomrule
\end{tabular}

\caption{Comparison of preference optimization methods on MedCalc-Bench using LLaMA3-8B-Instruct. All methods use the same SFT-Code initialization and preference dataset.}
\label{tab:pref-opt-ablation}
\end{table*}

\begin{table}[th]
    \centering
    \small
    \renewcommand\arraystretch{0.75}
    \begin{tabular}{lcccc}
        \toprule
        \multicolumn{1}{c}{\textbf{Loss}}&
        \multicolumn{1}{c}{\textbf{Reg.}}&
        \multicolumn{1}{c}{\textbf{Eq.}}&
        \multicolumn{1}{c}{\textbf{Rl.}}&
        \multicolumn{1}{c}{\textbf{Avg.}} \\
        \midrule

        $\mathcal{L}_{\text{SFT}}$&-&70.77&25.72&54.39 \\     
        \midrule
        $\mathcal{L}_{\text{DPO}}$&-&60.17&5.23&41.36\\    
                                  &0.5$\mathcal{L}_{\text{SFT}}$&62.82&8.66&43.13 \\
                                  &1.0$\mathcal{L}_{\text{SFT}}$&66.87&10.24&46.28 \\
        \midrule
        $\mathcal{L}_{\text{wDPO}}$&-&66.50&15.37&47.68 \\
                                  &0.5$\mathcal{L}_{\text{SFT}}$&67.32&20.27&52.17 \\
                                  &1.0$\mathcal{L}_{\text{SFT}}$&68.36&\textbf{29.40}&54.20 \\
                                  &adpt.&\textbf{72.26}&25.72&\textbf{55.34} \\

        \bottomrule
    \end{tabular}

    \caption{Results of regularization strategies on MedCalc-Bench using LLaMA3-8B-Instruct. ``adpt.' denotes the adaptive coefficient; \textbf{Eq.} and \textbf{Rl.} denote equation-based and rule-based tasks; ``--' indicates that no additional SFT regularization is used.}
    \label{tb:reg}
\end{table}

\subsection{Comparison with zero and few-shot}

We compare zero-shot and few-shot configurations to measure the effect of adding in-context demonstrations to the code-generation pipeline.
As shown in Table~\ref{tab:zero-few-shot-ablation-full}, increasing the number of demonstrations from zero to one and three raises the average score from 33.37 to 34.56 and 36.77, respectively. The improvement is consistent but modest relative to the gains from training-based code generation reported in Table~\ref{tb:main}, indicating that the main benefit arises from task-specific training rather than from additional inference-time examples.

\subsection{Comparison with Other Preference Optimization Approaches}

We compare wDPO with three representative preference optimization methods: rDPO~\citep{rdpo}, ROPO~\citep{ropo}, and WPO~\citep{wpo}.

All methods start from the same LLaMA3-8B-Instruct SFT-Code checkpoint and use the same MedCalc-Bench preference data. As shown in Table~\ref{tab:pref-opt-ablation}, their average accuracies are 50.84 for rDPO, 54.25 for ROPO, 52.33 for WPO, and 55.34 for wDPO. Thus, wDPO performs best among the compared preference methods and improves over the SFT-Code baseline of 54.39 by 0.95 percentage points; the other preference methods do not consistently exceed that baseline. wDPO also obtains the highest aggregate performance over the four equation-based categories, supporting the value of adaptive weighting for subtle numerical differences.

\section{Conclusion}

We introduce MedCode, a framework for improving LLM-based medical calculation through embedded code generation. Rather than producing a numerical answer directly, the fine-tuned model generates an executable script that instantiates the relevant medical calculator with values extracted from the input. A deterministic interpreter then performs the arithmetic and returns the result with its explanation and unit.

We construct an executable-code version of MedCalc-Bench and a synthetic dataset for longitudinal ICU calculation. In addition to SFT, we mine erroneous model-generated scripts to form preference pairs and introduce wDPO to weight these pairs according to their difficulty. Across the evaluated models and datasets, embedded code generation substantially improves calculation accuracy, while wDPO provides additional gains on selected tasks.

\section*{Limitations}

MedCode relies on predefined medical calculators and therefore may not generalize to formulas or clinical rules absent from the training data. The error-mining procedure is also limited by the failure modes produced by the SFT model and may underrepresent rare or clinically complex errors. Finally, our experiments focus on short, well-specified calculation tasks; performance in longer clinical workflows, across multiple patients, or under incomplete and conflicting records remains untested.

\section*{Ethics}

This work follows the ACL Code of Ethics. MedCalc-Bench is publicly available, and the ICU dataset consists of synthetic records manually authored and verified by experienced clinicians; it contains no real patient records or personally identifiable information. The datasets and models are intended for research on calculation reliability and have not been validated for autonomous clinical decision-making.

\bibliography{cite}

\appendix

\section{Coding Scripts Design details for MedCalc-based Dataset}
\label{app:eq-implmentations}

Table~\ref{app:code-example} presents representative embedded scripts. We use distinct script designs for equation-based and rule-based calculators.

\paragraph{Equation-based Tasks}

Equation-based tasks instantiate predefined medical formulas. Single-step calculators are represented by compact expressions, whereas multi-step calculators expose intermediate quantities explicitly. Dosage-conversion scripts define the required conversion constants, and date-based calculators use standard operations from \texttt{datetime}.

\paragraph{Rule-based Tasks}

Rule-based tasks evaluate calculator criteria against information in the patient note. The script accumulates a score by testing numerical ranges and categorical findings. Numerical inputs are handled with explicit conditional expressions, while categorical variables are assigned from the model interpretation of the clinical description.

We manually verify and execute all script templates. The resulting dataset contains 9,766 training examples and 1,048 test examples, covering 41 calculators grouped into four equation-based and three rule-based task types. Appendix~\ref{app:code-execution} describes the execution procedure.

\begin{table*}[th]
    \centering
    \tiny
    \setlength{\tabcolsep}{0.7mm} 
    \renewcommand\arraystretch{1.0}
    \begin{tabular}{lll}
        \toprule 
        \makecell[c]{\footnotesize\textbf{Category}} & &\makecell[c]{\footnotesize\textbf{Embedded Coding Script Design \textbf{(Explanation printings omitted)}}} \\
        \midrule
        \multirow{50}{*}{\makecell[l]{\footnotesize\textbf{Equation-based}}} & 
        \makecell[l]{{\footnotesize Single-step} } &
        {\footnotesize
        \begin{tabular}[t]{@{}l@{}}
            \# BMI = weight / (height * height)\\
            bmi = 76.0 / ((183.0 / 100) * (183.0 / 100))\\
            bmi = round(bmi, 3)\\
            print('The Body Mass Index (BMI) is ', bmi, 'kg/m²')
        \end{tabular}
        } \\
        \cmidrule(lr){2-3}
        & {\footnotesize Multi-step} &
        {\footnotesize
        \begin{tabular}[t]{@{}l@{}}
            \# GFR = 175 * creatinine\^{}(-1.154) * age\^{}(-0.203) * race\_coefficient * gender\_coefficient\\
            race\_coefficient = 1.212 if 'Black' == 'Black' else 1.0\\
            gender\_coefficient = 0.742 if 'Male' == 'Female' else 1.0\\
            gfr = 175 * 10.6**(-1.154) * 49**(-0.203) * race\_coefficient * gender\_coefficient\\
            gfr = round(gfr, 3)\\
            print('The MDRD GFR is ', gfr, 'mL/min/ 1.73 m²')
        \end{tabular}
        } \\
        \cmidrule(lr){2-3}
        & {\footnotesize Conversion} &
        {\footnotesize
        \begin{tabular}[t]{@{}l@{}}
            \# converted dose = (from mass) * (to reletive mass) / (from reletive mass)\\
            reletive\_mass = \{'Betamethasone IV': 1, 'Cortisone PO': 33.33, 'Dexamethasone IV': 1, \\'Dexamethasone PO': 1,'Hydrocortisone IV': 26.67,'Hydrocortisone PO': 26.67,\\'MethylPrednisoLONE IV': 5.33,'MethylPrednisoLONE PO': 5.33,'PrednisoLONE PO': 6.67, \\'PredniSONE PO': 6.67, 'Triamcinolone IV': 5.33\}\\
            converted\_dose = \\\ 190.936 * reletive\_mass['MethylPrednisoLONE IV'] / reletive\_mass['Hydrocortisone PO']\\
            converted\_dose = round(converted\_dose, 3)\\
            print('The Steroid Conversion is ', converted\_dose, 'mg')
        \end{tabular}
        } \\
        \cmidrule(lr){2-3}
        & {\footnotesize Date} &
        {\footnotesize
        \begin{tabular}[t]{@{}l@{}}
            import datetime \\
            \# estimated due date = input\_date + 40 weeks \\
            future\_date = datetime.datetime.strptime('11/13/2020', '\%m/\%d/\%Y') + \\ datetime.timedelta(weeks=40) \\
            cycle\_length\_gap = (-1)\*\*(23 < 28) \* datetime.timedelta(days=abs(23 - 28))\\
            estimated\_due\_date = (future\_date + cycle\_length\_gap).strftime('\%m/\%d/\%Y')\\
            print('The Estimated Due Date is ', estimated\_due\_date, '')
        \end{tabular}
        } \\
        \cmidrule(lr){2-3}
        & {\footnotesize Duration} &
        {\footnotesize
        \begin{tabular}[t]{@{}l@{}}
            import datetime\\
            \# estimated gestational age =  the number of weeks and days apart today's date is from \\ the patient's last menstrual period date\\
            current\_date = datetime.datetime.strptime('09/01/2009', '\%m/\%d/\%Y')\\
            menstrual\_date = datetime.datetime.strptime('05/21/2009', '\%m/\%d/\%Y')\\
            weeks = abs(current\_date - menstrual\_date).days // 7 \\
            days = abs(current\_date - menstrual\_date).days \% 7\\
            print('The Estimated Gestational Age is', weeks, 'weeks', days, 'days','')
        \end{tabular}
        } \\
        
        \midrule
        \multirow{18}{*}{\makecell[l]{\footnotesize\textbf{Rule-based}}} & 
        \makecell[l]{{\footnotesize Single-sym.}} &
        {\footnotesize
        \begin{tabular}[t]{@{}l@{}}
            \# Centor Score for Strep Pharyngitis criteria:\\
            \# age, temperature, tonsil swelling, lymph nodes\\
            \# age thresholds: (45, inf), (3, 15); temperature > 38°C\\
            score = 0\\
            score += -1 * (45 <= 48 < float('inf')) + 1 * (3 <= 48 < 15)\\
            score += 1 * (38 < ((99.0 - 32) * 5/9) < float('inf'))\\
            has\_cough\_absent = False; score += 1 if has\_cough\_absent else 0\\
            has\_tender\_lymph\_nodes = False; score += 1 if has\_tender\_lymph\_nodes else 0\\
            has\_exudate\_swelling\_tonsils = False; score += 1 if has\_exudate\_swelling\_tonsils else 0\\
            print('The Centor Score (Modified/McIsaac) for Strep Pharyngitis is', score)
        \end{tabular}
        } \\
        \cmidrule(lr){2-3}
        & {\footnotesize Multi-sym.} &
        {\footnotesize
        \begin{tabular}[t]{@{}l@{}}
            \# PERC Rule for Pulmonary Embolism criteria: age, heart\_rate, ...
            \# age thresholds are: (50, 'inf')\\ 
            ...
            \# The patient has Hormone use or not
            score = 0 \\
            score += 1 * (50 <= 73 < float('inf'))\\
            score += 1 * (100 <= 92.0 < float('inf'))\\
            score += 1 * (0 <= 98.0 < 95)\\
            has\_unilateral\_leg\_swelling = False; score += 1 if has\_unilateral\_leg\_swelling else 0\\
            has\_hemoptysis = False; score += 1 if has\_hemoptysis else 0\\
            has\_recent\_surgery\_or\_trauma = True; score += 1 if has\_recent\_surgery\_or\_trauma else 0\\
            has\_pre\_pe = False; has\_pre\_dvt = False; score += 1 if (has\_pre\_pe or has\_pre\_dvt) else 0\\
            has\_hormonal\_use = True; score += 1 if has\_hormonal\_use else 0\\
            print('The PERC Rule for Pulmonary Embolism is', score)
        \end{tabular}
        } \\
        \bottomrule
    \end{tabular}

    \caption{Examples of embedded scripts for medical calculation. ``single-sym.' denotes a criterion containing one symptom, whereas ``multi-sym.' denotes a criterion combining multiple symptoms (e.g., $\text{has\_pre\_pe or has\_pre\_dvt}$). Explanatory output statements are omitted for brevity.}
    \label{app:code-example}
\end{table*}

\section{Hyperparameter and Prompts Settings}
\label{app:prompt}

We follow the inference configuration released with MedCalc-Bench and use LLaMA-Factory defaults as the basis for SFT and preference training.

\paragraph{Data augmentation hyperparameters} We set $k=1{,}000$ to balance inference cost and the diversity of mined errors. Generating scripts over multiple temperatures produces approximately 1,500 preference pairs.

\paragraph{Training hyperparameters} MedCalc-Bench evaluates prompted base models and does not specify SFT or preference-training hyperparameters. We therefore implement both stages with LLaMA-Factory and report the settings in Table~\ref{app:hyper-sft}.

\paragraph{Inference hyperparameters} We follow the inference settings in the public MedCalc-Bench implementation. Table~\ref{app:hyper-inference} lists the values used in our experiments.

\paragraph{Prompts} We adapt the code-generation prompts released with MedCalc-Bench. Box~\ref{fig:prompt1} presents the prompt templates.


\section{Case Study of Minor Error And Substantial Error}
\label{app:case-study-error}

Box~\ref{box:dpo-err} uses a GFR calculation to contrast minor and substantial script errors. A minor error changes a local element, such as a coefficient, and can therefore be difficult to distinguish from the reference implementation. A substantial error removes or adds an entire computation step and creates a larger structural difference. wDPO does not require explicit error-category labels; it assigns weights from the model probability gap for each response pair.

We manually categorize generated scripts before and after wDPO as containing \emph{minor} errors (incorrect extracted values, coefficients, or constants, with at most two such errors per script) or \emph{substantial} errors (missing or unnecessary steps, or execution failures). Table~\ref{tab:error-type-analysis} shows that the number of minor-error cases decreases from 360 to 312, while substantial-error cases increase from 87 to 144. The reduction of 48 minor-error cases is consistent with improved discrimination of local mistakes, but the simultaneous increase in substantial errors means that the table should not be interpreted as an unqualified reduction in total failures.

\section{Coding Execution Implementation}
\label{app:code-execution}

Code execution has two stages. \textbf{(1) Preprocessing}: Following MedCalc-Bench, we use regular expressions to remove redundant whitespace and line breaks. \textbf{(2) Python execution}: We execute the generated script with \texttt{exec()} in the controlled evaluation environment and capture its standard output. Calculator descriptions remain in comments, while executable statements perform the calculation and print the response.

\section{Case Study}

Box~\ref{box:prompt3} presents a complete example in which a fine-tuned model generates and executes code for a medical calculation.

\newpage

\begin{NewBox}{Prompt Template 1: Direct Inference Prompt}
\label{fig:prompt1}
You are a helpful assistant for calculating a score for a given patient note. Please output answer only without any other text. Your output should only contain a JSON dict formatted as \{"answer": str(value which is the answer to the question)\}.
\\

Here is the patient note:

<Patient Note>
\\

Here is the task:

<Question>
\\

Please directly output the JSON dict formatted as \{{"answer": str(value which is the answer to the question)}\}:
\end{NewBox}

\begin{NewBox}{Prompt Template 2: CoT Inference Prompt (Zero-shot and One-shot)}
\label{fig:prompt2}
You are a helpful assistant for calculating a score for a given patient note. Please think step-by-step to solve the question and then generate the required score. Your output should only contain a JSON dict formatted as \\
\{{"step\_by\_step\_thinking": str(your\_step\_by\_step\_thinking\_procress\_to\_solve\_the\_question), "answer": str(short\_and\_direct\_answer\_of\_the\_question)}\}. 
\\

\textit{(Extra prompt for one-shot: start)}

Here is an example patient note:
\\

<Example Note>
\\

Here is an example task:
\\

<One Shot Question>
\\

Please directly output the JSON dict formatted as \\
\{{"step\_by\_step\_thinking": str(your\_step\_by\_step\_thinking\_procress\_to\_solve\_the\_question), "answer": str(value which is the answer to the question)}:{json.dumps(example\_output)}\}

\textit{(Extra prompt for one-shot: end)}\\

Here is the patient note:

<Patient Note>
\\

Here is the task:

<Question>
\\

Please directly output the JSON dict formatted as 

\{{"step\_by\_step\_thinking": str(your\_step\_by\_step\_thinking\_procress\_to\_solve\_the\_question), "answer": str(short\_and\_direct\_answer\_of\_the\_question)}\}. 

\end{NewBox}

\begin{NewBox}{Prompt Template 3: Coding Inference Prompt}
\label{box:prompt-template3}





You are a helpful assistant for calculating a score for a given patient note. Please think step-by-step to solve the question. Do not perform any computations yourself. Do not state a numerical answer. First write code for any equations you are using and then plug in values based on the patient note. Make sure the code prints all of its outputs. Your output should only contain a JSON dict formatted as \\
\{{"step\_by\_step\_thinking": str(your\_step\_by\_step\_thinking\_procress\_to\_solve\_the\_question), "code": str(medical\_calculation\_code)}\}. 
\\

\textit{(Extra prompt for one-shot: start)}

Here is an example patient note:
\\

<Example Note>
\\

Here is an example task:
\\

<One Shot Question>
\\

Please directly output the JSON dict formatted as \\
\{{"step\_by\_step\_thinking": str(your\_step\_by\_step\_thinking\_procress\_to\_solve\_the\_question), "code": str(example\_medical\_calculation\_code)}\}

\textit{(Extra prompt for one-shot: end)}\\

Here is the patient note:

<Patient Note>
\\

Here is the task:

<Question>
\\

Please directly output the JSON dict formatted as

\{{"step\_by\_step\_thinking": str(your\_step\_by\_step\_thinking\_procress\_to\_solve\_the\_question), "code": str(medical\_calculation\_code)}\}.

\end{NewBox}


\begin{table}[t]
\centering
\small
\setlength{\tabcolsep}{6pt}
\begin{tabular}{ll}
\toprule
\textbf{Hyperparameters} & \textbf{Value} \\
\midrule
batch size & 4 \\
learning rate & 5e-5 \\
epochs & 2.0 \\
optimizer & AdamW \\
cutoff length & 2048 \\
data type & bfloat16 \\
preprocessing workers & 24 \\
learning rate scheduler & cosine \\
\bottomrule
\end{tabular}
\caption{Supervised fine-tuning and preference training hyperparameters.}
\label{app:hyper-sft}
\end{table}

\begin{table}[t]
\centering
\small
\setlength{\tabcolsep}{6pt}
\begin{tabular}{ll}
\toprule
\textbf{Hyperparameters} & \textbf{Value} \\
\midrule
max length & 8192/ 4096\\
temperature & 0.0 \\
data type & float16 \\
do sample & False \\
truncation & True \\
\bottomrule
\end{tabular}

\caption{Inference hyperparameters. The maximum length is 8,192 for LLaMA3 and GPT-4 and 4,096 for GPT-3.5, following the MedCalc-Bench configuration.}
\label{app:hyper-inference}
\end{table}

\begin{table}[]
\centering
\small
\renewcommand\arraystretch{1.0}
\begin{tabular}{lcc}
\toprule
\textbf{Method} & \textbf{Minor} & \textbf{Substantial} \\
\midrule
w/o wDPO & 360 & 87 \\
w/ wDPO  & 312 & 144 \\
\bottomrule
\end{tabular}

\caption{Coding error types before and after wDPO. Minor errors comprise incorrect extracted values, coefficients, or constants, with at most two such errors per script. Substantial errors comprise missing or unnecessary steps and execution failures.}
\label{tab:error-type-analysis}
\end{table}


    

\begin{NewBox}{Case Study of Minor Error And Substantial Error on MDRD GFR Coding}\label{box:dpo-err}
\textbf{Correct:}

\begin{verbatim}
equation = 'GFR=175*creatinine^(-1.154)*age^(-0.203)*race_coeff *gender_coeff'
print('Calculation equation is ', equation)
race_coeff=1.212 if 'White' == 'Black' else 1.0
print('Patient race is White, race coefficient is ', race_coeff)
gender_coeff=0.742 if 'Male' == 'Female' else 1.0
print('Patien gender is Male, gender coefficient is ', gender_coeff)
print('Patient creatinine is 1.4 mg/dL')
print('Patient age is 73 years old')
gfr=175*1.4**(-1.154)*73**(-0.203)*race_coeff*gender_coeff
print('The MDRD GFR is', round(gfr, 3), 'mL/min/1.73 m²')
\end{verbatim}
\medskip

\textbf{Wrong-minor:} \texttt{gender\_coeff} value is wrong.

\begin{verbatim}
equation = 'GFR=175*creatinine^(-1.154)*age^(-0.203)*race_coeff *gender_coeff'
print('Calculation equation is ', equation)
race_coeff=1.212 if 'White' == 'Black' else 1.0
print('Patient race is White, race coefficient is ', race_coeff)
gender_coeff=0.942 if 'Male' == 'Female' else 1.0
print('Patien gender is Male, gender coefficient is ', gender_coeff)
print('Patient creatinine is 1.4 mg/dL')
print('Patient age is 73 years old')
gfr=175*1.4**(-1.154)*73**(-0.203)*race_coeff*gender_coeff
print('The MDRD GFR is', round(gfr, 3), 'mL/min/1.73 m²')
\end{verbatim}
\medskip

\textbf{Wrong-substantial:} missing definition of \texttt{gender\_coefficient =
0.742 if 'Male' == 'Female' else 1.0}.)

\begin{verbatim}
equation = 'GFR=175*creatinine^(-1.154)*age^(-0.203)*race_coeff'
print('Calculation equation is ', equation)
race_coeff=1.212 if 'White' == 'Black' else 1.0
print('Patient race is White, race coefficient is ', race_coeff)
print('Patient creatinine is 1.4 mg/dL')
print('Patient age is 73 years old')
gfr=175*1.4**(-1.154)*73**(-0.203)*race_coeff
print('The MDRD GFR is', round(gfr, 3), 'mL/min/1.73 m²')
\end{verbatim}
\end{NewBox}

\begin{NewBox}{Case Study: Glomerular Filtration Rate (GFR)}\label{box:prompt3}
\textbf{Instruction}: 




You are a helpful assistant for calculating a score for a given patient note. Please think step-by-step to solve the question. Do not perform any computations yourself. Do not state a numerical answer. First write code for any equations you are using and then plug in values based on the patient note. Make sure the code prints all of its outputs. Your output should only contain a JSON dict formatted as \\
\{{"step\_by\_step\_thinking": str(your\_step\_by\_step\_thinking\_procress\_to\_solve\_the\_question), "code": str(medical\_calculation\_code)}\}. 
\\

\textit{...(Extra prompt for one-shot setting)}
\\

Here is the patient note:

A 73-year-old Caucasian male presented with longstanding chronic kidney disease (CKD) G2 related to arterial hypertension stage II. Past medical history: prostate cancer and monoclonal gammopathy of undetermined significance (MGUS) IgG k, rosacea dermatitis, right hand arthritis. Laboratory tests () showed normal complete blood count (CBC), normal serum calcium value, creatinine 1.4 mg/dL, B type natriuretic peptide (BNP) 161 pg/mL, proteinuria 946 mg/24 h, ...\textit{(Detailed Patient Notes)}...
moreover, it could be not uncommon to discover a plasma cell disorder in patients with SAA amyloid, given the increasing incidence of both conditions with chronic inflammatory disease.
\\

Here is the task:

Using the MDRD GFR Equation, what is the patient's Glomerular Filtration Rate (GFR) in terms of mL/min/1.73 m²? If the patient is black, please use the MDRD GFR Equation for Blacks. You should use the patient's medical values and health status when they were first admitted to the hospital prior to any treatment.
\\

Please directly output the JSON dict formatted a

\{{"step\_by\_step\_thinking": str(your\_step\_by\_step\_thinking\_procress\_to\_solve\_the\_question), "code": str(medical\_calculation\_code)}\}. 
\\

\textbf{Example Embedded Coding Script (with values embedded):}
\begin{verbatim}
equation = 'GFR=175*creatinine^(-1.154)*age^(-0.203)*race_coeff *gender_coeff'
print('Calculation equation is ', equation)
race_coeff=1.212 if 'White' == 'Black' else 1.0
print('Patient race is White, race coefficient is ', race_coeff)
gender_coeff=0.742 if 'Male' == 'Female' else 1.0
print('Patien gender is Male, gender coefficient is ', gender_coeff)
print('Patient creatinine is 1.4 mg/dL')
print('Patient age is 73 years old')
gfr=175*1.4**(-1.154)*73**(-0.203)*race_coeff*gender_coeff
print('The MDRD GFR is', round(gfr, 3), 'mL/min/1.73 m²')
\end{verbatim}

\textbf{Execution Result:}
\begin{verbatim}
Calculation equation is  
GFR=175*creatinine^(-1.154)*age^(-0.203)* race_coeff *gender_coeff
Patient race is White, race coefficient is  1.0
Patient gender is Male, gender coefficient is  1.0
Patient creatinine is 1.4 mg/dL
Patient age is 73 years old
The MDRD GFR is 49.677 mL/min/1.73 m²
\end{verbatim}
\end{NewBox}


\begin{NewBox}{ICU Dataset: A translated data sample in ICU dataset}\label{box:icu-dataset-new}
\textbf{Instruction:}

Based on the following known information, respond concisely and professionally to the physician’s inquiry about the patient's condition. The answer should be in Chinese. Do not fabricate any numerical date values not present in the known content.
\\

\textbf{EHR:}

Known information about the patient's condition:

Vital sign Non-invasive systolic pressure 2024-09-14 00:00:00 123.2mmHg

Vital sign Invasive systolic pressure 2024-09-14 00:21:00 101.4mmHg

Daily vital sign Invasive blood pressure 2024-09-14 00:32:00 153.6mmHg

Vital sign Non-invasive diastolic pressure 2024-09-14 01:19:00 57.2mmHg

Vital sign Invasive diastolic pressure 2024-09-14 01:33:00 73.3mmHg

Daily vital sign Invasive blood pressure 2024-09-14 01:37:00 146.9mmHg

Daily vital sign Non-invasive blood pressure 2024-09-14 01:54:00 91.6mmHg

Vital sign Non-invasive systolic pressure 2024-09-14 02:26:00 117.4mmHg

Please answer the clinician’s inquiry about the patient's condition: List the cases where the patient’s blood pressure is $\leq$ 137.2mmHg and how many times this condition is met.
\\

\textbf{Example Coding:}
\begin{verbatim}
print('Vital sign Non-invasive systolic pressure 2024-09-14 00:00:00 123.2mmHg' \
if 123.2 <= 137.2 else '')
print('Vital sign Invasive systolic pressure 2024-09-14 00:21:00 101.4mmHg' \
if 101.4 <= 137.2 else '')
print('Daily vital sign Invasive blood pressure 2024-09-14 00:32:00 153.6mmHg'\
if 153.6 <= 137.2 else '')
print('Vital sign Non-invasive diastolic pressure 2024-09-14 01:19:00 57.2mmHg'\
if 57.2 <= 137.2 else '')
print('Vital sign Invasive diastolic pressure 2024-09-14 01:33:00 73.3mmHg'\
if 73.3 <= 137.2 else '')
print('Daily vital sign Invasive blood pressure 2024-09-14 01:37:00 146.9mmHg'
\if 146.9 <= 137.2 else '')
print('Daily vital sign Non-invasive blood pressure 2024-09-14 01:54:00 91.6mmHg'\
if 91.6 <= 137.2 else '')
print('Vital sign Non-invasive systolic pressure 2024-09-14 02:26:00 117.4mmHg'\
if 117.4 <= 137.2 else '')
print('Total', sum(1 for data in [123.2, 101.4, 153.6, 57.2, 73.3, 146.9, \
91.6, 117.4] if data <= 137.2), 'instances of blood pressure<137.2mmHg')
\end{verbatim}
\medskip

\textbf{Example Outputs:}
\begin{verbatim}
Vital sign Non-invasive systolic pressure 2024-09-14 00:00:00 123.2mmHg
Vital sign Invasive systolic pressure 2024-09-14 00:21:00 101.4mmHg
Vital sign Non-invasive diastolic pressure 2024-09-14 01:19:00 57.2mmHg
Vital sign Invasive diastolic pressure 2024-09-14 01:33:00 73.3mmHg
Daily vital sign Non-invasive blood pressure 2024-09-14 01:54:00 91.6mmHg
Vital sign Non-invasive systolic pressure 2024-09-14 02:26:00 117.4mmHg
Total 6 instances of blood pressure < 137.2mmHg
\end{verbatim}

\end{NewBox}

\begin{table*}[t]
\centering
\small
\renewcommand\arraystretch{1.0}
\setlength{\tabcolsep}{6pt}
\begin{tabular}{@{} l p{7.2cm} l l @{}}
\toprule
\textbf{Category} & \textbf{Metric} & \textbf{Unit} & \textbf{Range} \\
\midrule
\multirow[c]{2}{*}{Oxygen Saturation} 
& Blood gas analysis oxygen saturation & \% & [80, 100] \\
& Vital sign oxygen saturation (SpO$_2$) & \% & [80, 100] \\
\addlinespace

\multirow[c]{6}{*}{Blood Pressure} 
& Vital sign invasive diastolic pressure & mmHg & [40, 90] \\
& Vital sign non-invasive diastolic pressure & mmHg & [40, 90] \\
& Vital sign invasive systolic pressure & mmHg & [90, 140] \\
& Vital sign non-invasive systolic pressure & mmHg & [90, 140] \\
& Same-day vital sign invasive blood pressure & mmHg & [70, 160] \\
& Same-day vital sign non-invasive blood pressure & mmHg & [70, 160] \\
\addlinespace

\multirow[c]{2}{*}{CO$_2$ Partial Pressure} 
& Blood gas analysis partial pressure of CO$_2$ (pCO$_2$) & mmHg & [35, 45] \\
& Blood gas analysis temperature-corrected partial pressure of CO$_2$ (pCO$_2$) & mmHg & [35, 45] \\
\addlinespace

\multirow[c]{1}{*}{Bilirubin} 
& Liver function direct bilirubin & $\mu$mol/L & [0, 6.8] \\
\addlinespace

\multirow[c]{2}{*}{Blood Calcium Concentration} 
& Electrolyte test ionized calcium & mmol/L & [1.1, 1.3] \\
& Electrolyte test total calcium & mmol/L & [2.1, 2.6] \\
\addlinespace

\multirow[c]{2}{*}{Serum Calcium} 
& Electrolyte test ionized calcium & mmol/L & [1.1, 1.3] \\
& Electrolyte test total calcium & mmol/L & [2.1, 2.6] \\
\addlinespace

\multirow[c]{1}{*}{Blood Potassium} 
& Electrolyte test potassium & mmol/L & [3.5, 5.0] \\
\bottomrule
\end{tabular}
\caption{Categories, metrics, units, and clinically plausible ranges in our ICU dataset.}
\label{tab:icu-metrics}
\end{table*}

\end{document}